\pdfoutput=1
\PassOptionsToPackage{prologue,dvipsnames}{xcolor}
\documentclass[11pt]{article}

\usepackage{acl}

\usepackage{times}
\usepackage{latexsym}
\usepackage{hyperref}
\usepackage{pgfplots}

\usepackage{tikz}
\usetikzlibrary{positioning, fit}

\usepackage[T1]{fontenc}

\usepackage[utf8]{inputenc}

\usepackage{microtype}

\usepackage{inconsolata}

\usepackage{graphicx}
\usepackage{multirow}
\usepackage{booktabs}
\title{BeaverTalk: Oregon State University's IWSLT 2025 \\ Simultaneous Speech Translation System}

\author{Matthew Raffel, Victor Agostinelli, \and Lizhong Chen \\
       Oregon State University \\
\texttt{\{raffelm, agostinv, chenliz\}@oregonstate.edu}
}

\begin{document}
\maketitle
\begin{abstract}
This paper discusses the construction, fine-tuning, and deployment of BeaverTalk\footnote{Our fine-tuning and evaluation code is available at \url{https://github.com/OSU-STARLAB/BeaverTalk}}, a cascaded system for speech-to-text translation as part of the IWSLT 2025 simultaneous translation task. The system architecture employs a VAD segmenter for breaking a speech stream into segments, Whisper Large V2  for automatic speech recognition (ASR), and Gemma 3 12B for simultaneous translation.  Regarding the simultaneous translation LLM, it is fine-tuned via low-rank adaptors (LoRAs) for a conversational prompting strategy that leverages a single prior-sentence memory bank from the source language as context.  The cascaded system participated in the English$\rightarrow$German and English$\rightarrow$Chinese language directions for both the low and high latency regimes. In particular, on the English$\rightarrow$German task, the system achieves a BLEU of 24.64 and 27.83 at a StreamLAAL of 1837.86 and 3343.73, respectively.  Then, on the English$\rightarrow$Chinese task, the system achieves a BLEU of 34.07 and 37.23 at a StreamLAAL of 2216.99 and 3521.35, respectively.
\end{abstract}

\section{Introduction}
This paper covers Oregon State University's simultaneous translation system, BeaverTalk, for IWSLT 2025. The system constructed takes in a speech stream input and outputs text translation in a cascaded manner for two language pairs, those being English$\rightarrow$German (en$\rightarrow$de) and English$\rightarrow$Chinese (en$\rightarrow$zh). Unique to IWSLT 2025's simultaneous translation task \cite{abdulmumin-etal-2025-findings}, this system generates translation for unsegmented audio. Architecture-wise, the system includes a VAD speech segmenter \citep{SileroVAD}, breaking a speech stream into segments, Whisper Large V2 \citep{radford2022whisper} performing automatic speech recognition (ASR), and a fine-tuned Gemma 3 12B model \citep{gemmateam2025gemma3technicalreport} that performs context-aware conversational prompting to generate a simultaneous translation.

The simultaneous translation portion of this cascaded system is fine-tuned on OpenSubtitles v2018 \citep{lison-etal-2018-opensubtitles2018} across both language pairs. Given the unsegmented source for this task, leveraging additional context is possible and likely to improve results, based on prior work \citep{papi2024streamattdirectstreamingspeechtotext}. As such, our system utilizes a single-sentence memory bank for the source language as context. This memory bank required modifying the typical conversational prompting structure for simultaneous translation\cite{wang2024conversationalsimulmtefficientsimultaneous}.

Although a fine-tuned Gemma 3 12B leveraging conversational prompting is a powerful model for simultaneous translation, its application in a cascaded architecture suffers from typical issues of error propagation \citep{tran-etal-2022-joint, zhou2024prosody}.  As such, maximizing the capabilities of a powerful simultaneous translation LLM requires minimizing these errors in the preceding steps, consisting of the VAD segmenter and Whisper ASR model.  As such, we conduct an extensive inference time hyperparameter search aimed at minimizing error propagation.  From the joint contributions of our cascaded simultaneous translation system and minimization of error propagation, we achieve impressive results on the ACL 60/60 development set.  For example, on the English$\rightarrow$German language pair, our cascaded system achieves a BLEU of 24.64 and 27.83 at a streamLAAL of 1837.86 and 3343.73.  Furthermore, on the English$\rightarrow$Chinese task, our system achieves a BLEU of 34.07 and 37.23 at a streamLAAL of 2216.99 and 3521.35.
\begin{figure*}[t]
    \centering
    \includegraphics[width=\linewidth]{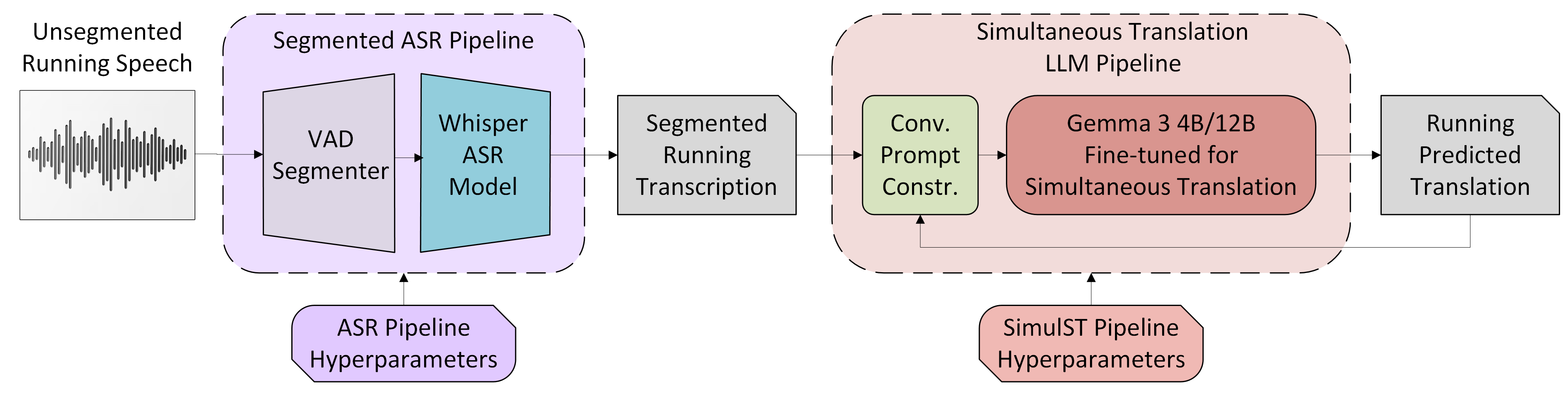}
    \caption{Depiction of the cascaded system described in this technical paper. Unsegmented source audio is taken in and fed into an ASR pipeline that segments the audio and then transcribes it into a hard text data modality. This segmented, running transcription is then fed into a simultaneous translation pipeline powered by Gemma 3. The transcription and current translation are fed into a conversational prompt constructor, adapted from prior work.}
    \label{fig:system_description}
\end{figure*}

\section{Task Description}
Simultaneous translation, generally speaking, is the process of taking in some source context and making translation decisions to another language in a manner that does not rely on that source context being complete. For example, typical neural machine translation (NMT) might act on a sentence-to-sentence basis, taking in a source sentence and outputting a target sentence. In comparison, a simultaneous translation must balance the lagging factor of output translations (i.e., the time it takes from a piece of source context to a corresponding piece of the output translation) with translation quality, making translation decisions with only partial context.

As previously mentioned, the IWSLT 2025 simultaneous translation task \cite{abdulmumin-etal-2025-findings} is fundamentally a speech-to-text task with two tracks governing what systems participants are expected to build: text-to-text where participants only construct a simultaneous agent for text data and prepend this system with an ASR model and speech-to-text where the simultaneous system takes in raw speech and outputs target translations in text without the need for a conversion to a text data modality. Our constructed system targets the text-to-text track, and since it is applied to the English$\rightarrow$German (en$\rightarrow$de) and English$\rightarrow$Chinese (en$\rightarrow$zh) language directions, it is restricted to predefined high and low latency regimes specified by the task. These two latency regimes, as specified below, are governed by non-computationally aware StreamLAAL in seconds (s):

\begin{itemize}
    \item en$\rightarrow$de: 0-2s (low), 2-4s (high);
    \item en$\rightarrow$zh: 0-2.5s (low), 2.5-4s (high).
\end{itemize}

The required development set for en$\rightarrow$de and en$\rightarrow$zh is ACL 60/60. A blind test set is employed for final evaluations.

\section{BeaverTalk: A System Description}
\label{sec:system_description}
Our simultaneous translation system consists of a cascaded architecture, which is divided into a VAD segmenter utilizing a Silero VAD model \citep{SileroVAD}, Whisper Large V2 \cite{radford2022whisper}, and a fine-tuned Gemma 3 \citep{gemmateam2025gemma3technicalreport} for simultaneous translation. In the system, Gemma 3 was fine-tuned for a conversational prompting strategy \cite{wang2024conversationalsimulmtefficientsimultaneous}, which is designed to mimic a streaming setting.  Our complete system is provided in Figure \ref{fig:system_description}.  

Our choice of a cascaded architecture rather than an end-to-end system hinges on our desire to (1) leverage the language modeling capabilities of an LLM to overcome contextual obstacles faced during simultaneous translation, (2) take advantage of an LLMs context understanding capabilities to harness prior sentence context in a stream of data, and (3) benchmark the fine-tuning and multilingual capabilities of the recent Gemma 3 \cite{gemmateam2025gemma3technicalreport}. To provide a deeper understanding of our designed system, we will first explain the fine-tuning approach for our translation LLM to enable conversational prompting, followed by a deeper explanation of our cascaded translation system.

\begin{figure*}[t]
  \centering
  \resizebox{\textwidth}{!}{%
    \begin{tikzpicture}[
        node distance=1mm and 3mm,
        box/.style={
          draw,
          rectangle,
          rounded corners,
          align=left,
          font=\small,
          inner sep=6pt
        },
        outerbox/.style={
          draw=black,
          thick,
          rounded corners,
          inner sep=8pt
        }
      ]
      \node[box, fill=gray!10] (P) {%
        Use the following sentence as context: 
        \textcolor{blue}{$c_1,\dots,c_{|C|}$}\\
        Now translate the following sentence from X to Y Assistant:
      };
      \node[box, fill=green!10, right=of P] (A) {%
        \texttt{<s><t>}%
        \textcolor{red}{$s_1,\dots,s_i$}%
        \texttt{</t>}%
        \textcolor{teal}{$t_1,\dots,t_j$}%
        \texttt{</s>}%
      };
      \node[box, fill=white, right=of A] (E) {$\dots$};
      \node[box, fill=green!10, right=of E] (B) {%
        \texttt{<s><t>}%
        \textcolor{red}{$s_k,\dots,s_{|S|}$}%
        \texttt{</t>}%
        \textcolor{teal}{$t_l,\dots,t_{|T|}$}%
        \texttt{</s>}%
      };

      \node[outerbox, fit=(P) (A) (E) (B)] {};
    \end{tikzpicture}%
  }
  \caption{An example conversational prompt for translating for source language X to target language Y using source and target sequences $S = [s_1, s_2, ..., s_{|S|}]$ and $T=[t_1, t_2, ..., t_{|T|}]$ with context $C=[c_1, c_2, ..., c_{|C|}]$.}
  \label{fig:prompt-fullwidth}
\end{figure*}
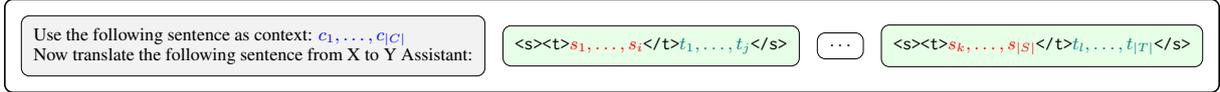

\subsection{SFT Conversational Prompting}
\label{sec:SFTconv}
We conduct supervised fine-tuning (SFT) for our Gemma 3 LLM for translation using a conversational prompting strategy that leverages a prior sentence memory bank as context.  The prompting strategy is designed whereby provided a source sequence $S = [s_1, s_2, ..., s_{|S|}]$ and a target sequence $T=[t_1, t_2, ..., t_{|T|}]$ the prompt will interleave subsequences from $S$ and $T$ leveraging delimiting tokens to separate the subsequences.  Now suppose we had prior sentence context $C=[c_1, c_2, ..., c_{|C|}]$, then an example of a conversational prompt constructed from these components is provided in Figure \ref{fig:prompt-fullwidth}.

Aside from the prior sentence memory bank, which we inject into our prompt, our conversational prompting follows a similar implementation to \citet{wang2024conversationalsimulmtefficientsimultaneous}.  The approach for generating this conversational prompting (the green region in Figure \ref{fig:prompt-fullwidth}) can be broken into the following three steps:
\begin{enumerate}
    \item Generate the alignments between words in the source and the target sequences. Unlike \citet{wang2024conversationalsimulmtefficientsimultaneous}, which uses fast-align \citep{dyer-etal-2013-simple}, we use the Itermax method from the SimAlign toolkit leveraging XLM-RoBERTa base to align words due to their work reporting better alignments \citep{jalili-sabet-etal-2020-simalign, DBLP:journals/corr/abs-1911-02116}.
    \item Segment the graph into subsequences such that all the word dependencies for each target subsequence are available in or before the respective source subsequence.  For example, assuming we did not perform step 3, in Figure \ref{fig:prompt-fullwidth} every word in the subsequence $t_1, ...t_j$ aligns with each word in $s_1, ..., s_i$.
    \item Merge and shift subsequences to break the ideal alignments.  Such a step is necessary to aid in making the LLM flexible for different variations of subsequences received during inference.
\end{enumerate}
Once the prompt is constructed, we fine-tune our LLM using a causal language modeling objective using cross-entropy loss.  We ensure that loss is only computed for tokens between the delimiting tokens \texttt{</t>}, not inclusive, and \texttt{</s>}, inclusive.  Suppose our conversational prompt possesses $K$ conversation intervals (ie. the number of times\texttt{<s>} appears in the prompt of Figure \ref{fig:prompt-fullwidth}), where the beginning and end of each conversation interval are at index $s_k$ and $e_{k}$.  Then we can represent such a loss objective with Equation \ref{eq:loss}.
\begin{equation}
\mathcal{L}
=
\sum_{k=1}^{K}
\sum_{i=s_{k}}^{e_{k}}
\log p_{\theta}\bigl(t_{i}\mid s_{<i}\bigr)
\label{eq:loss}
\end{equation}

The purpose of such a loss is to ensure that the model learns to predict \texttt{</s>} whenever it has insufficient context at inference.  In doing so our LLM learns a portion of the decision policy in conjunction with the translation objective. 

\subsection{Streaming Cascaded SimulST System}
As we leverage a cascaded architecture we will break our explanation into (1) the Segmented ASR Pipeline, the part responsible for segmenting and transcribing a speech stream (shown in the left half of Figure \ref{fig:system_description}), and (2) the Simultaneous Translation LLM Pipeline, the part responsible for translating the transcribed speech (shown in the right half of Figure \ref{fig:system_description}). 
\subsubsection{Segmented ASR Pipeline}
The first part of our Segmented ASR Pipeline is the VAD segmenter. 
As previously mentioned, it segments a speech signal.  This segmentation is based on (1) the maximum segment duration, (2) the maximum unvoiced duration, and (3) the voice probability threshold.  As the name implies, the maximum segment duration determines the maximum length of a valid segment.  If the segment duration exceeds the maximum segment duration, it is cut. The maximum unvoiced duration and the voice probability threshold alternatively rely on one another to determine when to segment the speech input prior to reaching the maximum segment duration.  The first part of this second facet of segmentation begins with the Silero VAD model \cite{SileroVAD}.  This VAD model outputs a probability score of a specific sample containing audio of a speaker's voice.  If the score falls below the voice probability threshold, it is determined that there is currently no voice from the speaker.  When the score has been below the probability threshold for longer than the maximum unvoiced duration, the speech is segmented. Such a condition would be ideally met between pauses in speech or at the end of a sentence.

The Whisper ASR model \citep{radford2022whisper} interacts with the VAD segmenter by receiving the segmented audio inputs. The Whisper portion of the Segmented ASR Pipeline is designed to have Whisper transcribe the audio input using a stable transcription policy, leveraging a context mechanism.  The stable transcription policy followed aims to create consistent, accurate transcriptions.  It works by committing a transcription to a stable transcription buffer once it repeats a transcription for a given audio interval.  For example, if on the first interval Whisper transcribes the sequence $s_1, s_2, s_3, s_4$ and then on the second interval it transcribes $s_1, s_2, s_3, s_4', s_5, s_6$, only the $s_1, s_2, s_3$ will be committed to the stable transcription.  Once committed as a stable transcription, it becomes available to the Simultaneous Translation LLM Pipeline for translation.  To further improve transcription quality, Whisper is also provided with additional context from a context buffer.  The context buffer is designed to provide Whisper with the transcript from the previous segment.  However, if the previous segment exceeds the cutoff threshold, the number of context words is limited to be equal to the cutoff threshold.

\subsubsection{Simultaneous Translation LLM Pipeline}
As previously explained in Section \ref{sec:SFTconv}, our Gemma 3 \cite{gemmateam2025gemma3technicalreport} based simultaneous translation model follows a conversational prompting strategy utilizing a prior sentence memory bank as context.  It is designed to firstly place the running stream of transcription chunks from the Segmented ASR Pipeline into a buffer upon receipt. Such a buffer will retain already translated portions of the transcript so long as the sentence these translated portions are associated with has yet to be completed.  

Once the buffer has been extended, it is passed to a Spacy sentence tokenizer, which splits the buffer of words into sentences. Upon splitting the buffer into sentences, the pipeline will enter a translation generation loop, where a translation action will occur if one of two conditions is met.  These conditions consist of (1) the length of untranslated words in the buffer has exceeded a prespecified minimum chunk size or (2) the sentence tokenizer has split the buffer into more than 1 sentence.  

Once a translation action is triggered, the first step is to construct a conversational prompt.  The conversational prompt is constructed identically to the one in Figure \ref{fig:prompt-fullwidth} by appending the new source subsequence from the oldest sentence in the buffer after a \texttt{<t>} delimiter following the previous translation action conversational prompt.  In order to ensure the model understands it is a conversation phase after the source sequence, the \texttt{</t>} delimiter is appended.  The new source subsequence length is equal to the untranslated word count present in the current sentence.  We require such a condition as allowing for multiple sentences in the source subsequence would deviate from the fine-tuning setting, where only a single sentence was allowed at any time in the conversational prompt (a restriction by the dataset).  Once constructed, the prompt is provided to the LLM to produce a translation until it outputs the delimiting token \texttt{</s>}.  The output translation is added to a running translation to be reused in prompt construction for subsequent translation phases.

Upon completing the translation, if the sentence tokenizer determined there was more than a single sentence in the buffer, it would signify that the current sentence had completed translation.  As such, the translated sentence transcript is cached to be used as context for the subsequent sentence.  Additionally, its contents would be removed from the transcription buffer.  Further translation phases would occur if conditions (1) and (2) were once again met.  

\section{Experimental Setup}
\begin{table*}[t]
  \centering
  \caption{Comparison table for simultaneous translation experiments, organized by language pair and model size.}
  \label{tab:comparisons}
  \begin{tabular}{ c | c | c | c | c }
    \hline
    Language Pair 
      & Model Size 
      & Latency Regime 
      & BLEU $\uparrow$
      & StreamLAAL $\downarrow$ \\
    \hline
    \multirow{4}{*}{en--de}
      & \multirow{2}{*}{4B}
          & low  & 23.64 & 1958.22 \\ \cline{3-5}
      &                    & high & 25.22 & 3503.46 \\ \cline{2-5}
      & \multirow{2}{*}{12B}
          & low  & 24.64 & 1837.86 \\ \cline{3-5}
      &                    & high & 27.83 & 3343.73 \\ \hline
    \multirow{4}{*}{en--zh}
      & \multirow{2}{*}{4B}
          & low  & 32.81 & 2249.32 \\ \cline{3-5}
      &                    & high & 34.62 & 3190.68 \\ \cline{2-5}
      & \multirow{2}{*}{12B}
          & low  & 34.07 & 2216.99 \\ \cline{3-5}
      &                    & high & 37.23 & 3521.35 \\ \hline
  \end{tabular}
\end{table*}
The dataset of choice for fine-tuning is OpenSubtitles v2018. This corpus is particularly noisy (e.g. some Chinese translations are almost entirely English, mismatched translations to transcriptions, etc.), rendering it difficult to achieve reasonable initial results. Given that, some cleaning of the dataset is required. This occurs in four steps, the third of which only occurs for the en$\rightarrow$zh dataset split:

\begin{enumerate}
    \item Filtering all samples on length such that the source and memory bank sequence are greater than or equal to 25 characters.  
    \item Filtering all samples with '...', '[', ']', '(', ')', or consisting of only capital letters and replacing '- ' with empty space.
    \item Filtering all samples in the en$\rightarrow$zh language split that contain English words in the target column.
    \item Filtering all remaining samples via CometKiwi \citep{rei2022cometkiwi} with a thresholding score of 0.6 to ensure semantic similarity between the transcription and the reference translation. This is meant to minimize the likelihood of a mismatched transcription and translation.
\end{enumerate}

The fine-tuning pipeline that employs the aforementioned dataset is based on frameworks for simultaneous translation with LLMs provided in prior work \citep{agostinelli-etal-2024-simul, raffel-etal-2024-simultaneous}, which were then adapted for unsegmented fine-tuning and evaluation. 

Fine-tuning occurred on Gemma 3 4B/12B via LoRAs with quantization \cite{hu2021loralowrankadaptationlarge, dettmers2023qlora}.  The LoRA adapters were applied to all attention projections and all the feed-forward network linear projections. We chose a LoRA $r$ of 64, a LoRA $\alpha$ of 16, and a LoRA dropout of 0.1.  Our quantization quantized to 4-bit floating point via NormalFloat with a compute data type of bfloat16.  We used the Paged 32-bit Adamw optimizer and an inverse sqrt learning rate scheduler with an effective batch size of 64, a learning rate of $2e^{-4}$, a weight decay of 0.1, a max gradient norm of 1, and a warm-up ratio of 0.03.

We evaluate the translation quality of our models using BLEU score with sacreBLEU \citep{post2018call}. The latency is reported using StreamLAAL \citep{papi2024streamattdirectstreamingspeechtotext}.  For the en$\rightarrow$de language direction, the latency and BLEU scores are reported at the word level using the 13a tokenizer.  Alternatively, for the en$\rightarrow$zh language direction, the latency and BLEU scores are reported at the character level.  
 
 Our fine-tuning and evaluation for the Gemma 12B models was conducted on an NVIDIA H200.  Alternatively, the Gemma 4B models were trained on a NVIDIA A40 and evaluated on a NVIDIA V100.

\section{Results}
\subsection{Inference Hyperparameter Tuning}
For our system, we tuned the maximum unvoiced duration (MUD), the voice probability threshold (VPT), the maximum segment duration (MSD), and the minimum chunk size (MCS).  We immediately found that a maximum unvoiced duration greater than 0.1 s would deteriorate performance, so we kept that constant for our experimentation.  Due to our 12B Gemma 3 model requiring an H200 to evaluate (a byproduct of memory requirements), we selected inference hyperparameters using a 4B Gemma 3 model, which could run on a V100.  Such a choice is feasible due to the cascaded structure of our architecture.  This is a byproduct of the maximum unvoiced duration, voice probability threshold, and maximum segment duration only influencing the quality of the transcriptions from the Segmented ASR Pipeline.  If the transcription related hyperparameters are tuned properly, the Gemma 3 translation model will perform better irrespective of the model size.  

We began our inference hyperparameter search by fixing our translation minimum chunk size to 3 words on the en$\rightarrow$de language pair and 5 words on the en$\rightarrow$zh language pair.  These minimum chunk sizes were chosen to accommodate the en$\rightarrow$de language pair, having a low latency cutoff of 2s, and the en$\rightarrow$zh language pair, having a low latency cutoff of 2.5s.  We then iteratively searched for the optimal maximum segment duration and voice probability threshold for both the low and high latency regimes.  We report these results in Tables \ref{tab:dehyper} and \ref{tab:zhhyper} for en$\rightarrow$de and en$\rightarrow$zh language pairs, respectively.

\begin{table}[h]
  \centering
  \caption{Simultaneous translation results for \textbf{en$\rightarrow$de} organized by voice probability threshold (VPT) and maximum segment duration (MSD) with a minimum chunk size of 3.}
  \label{tab:dehyper}
  \begin{tabular}{c|c|c|c}
    \hline
    VPT & MSD & BLEU $\uparrow$ & StreamLAAL $\downarrow$ \\
    \hline
    0.1 & 0.5 & 23.43 & 2079.14 \\
    0.1 & 1 & 24.86 & 2677.30 \\
    0.1 & 1.5 & 25.02 & 3114.15 \\
    \hline
    0.3 & 0.5 & 23.36 & 2047.62 \\
    0.3 & 1 & 25.01 & 2477.67 \\
    0.3 & 1.5 & 25.05 & 2877.81 \\
    \hline
    0.5 & 0.5 & 23.59 & 1940.58 \\
    0.5 & 1 & 24.96 & 2356.68 \\
    0.5 & 1.5 & 24.82 & 2804.01 \\
    \hline
  \end{tabular}
\end{table}

\begin{table}[h]
  \centering
  \caption{Simultaneous translation results for \textbf{en$\rightarrow$zh} organized by voice probability threshold (VPT) and maximum segment duration (MSD) with a minimum chunk size of 5.}
  \label{tab:zhhyper}
  \begin{tabular}{c|c|c|c}
    \hline
    VPT & MSD & BLEU $\uparrow$ & StreamLAAL $\downarrow$ \\
    \hline
    0.1 & 0.5 & 33.16 & 2294.95 \\
    0.1 & 1 & 33.57 & 2979.20 \\
    0.1 & 1.5 & 34.11 & 3382.83 \\
    \hline
    0.3 & 0.5 & 32.47 & 2307.56 \\
    0.3 & 1 & 33.36 & 2851.60 \\
    0.3 & 1.5 & 33.70 & 3206.28 \\
    \hline
    0.5 & 0.5 & 32.81 & 2249.32 \\
    0.5 & 1 & 33.31 & 2728.01 \\
    0.5 & 1.5 & 34.62 & 3190.68 \\
    \hline
  \end{tabular}
\end{table}

From observing, Table \ref{tab:dehyper} we can see for the low latency regime of the en$\rightarrow$de language pair the only selection of hyperparameters that fall below the 2s threshold is with a voice probability threshold of 0.5 and a maximum segment duration of 0.5s.  Alternatively, for the high latency regime, we select a voice probability threshold of 0.3 and a maximum segment duration of 1.  We chose the maximum segment duration of 1s rather than 1.5s due to the increase in StreamLAAL.   We report our final selected maximum unvoiced duration, voice probability threshold, and maximum segment duration in Table \ref{tab:final_en-de-hypers} for the en$\rightarrow$de language pair.
\begin{table}[h]
\centering
\caption{Inference hyperparameters for \textbf{en$\rightarrow$de}.}
\label{tab:final_en-de-hypers}
\begin{tabular}{c|c|c|c|c}
\hline
Latency & MUD     & VPT      & MSD  & MCS \\ \hline
low            & 0.1 & 0.5 & 0.5  & 3    \\
high           & 0.1  & 0.3 & 1  & 7    \\ \hline
\end{tabular}
\end{table}

On the high latency regime of the en$\rightarrow$zh from observing Table \ref{tab:zhhyper} we chose a voice probability threshold of 0.5 with a maximum segment duration of 1.5s due to the high BLEU achieved.  Then, for the low-latency regime, we chose a voice probability threshold of 0.5 and a maximum segment duration of 0.5 to align with our high-latency regime.  We report our final selected maximum unvoiced duration, voice probability threshold, and maximum segment duration in Table \ref{tab:final_en-zh-hypers} for the en$\rightarrow$zh language pair.

\begin{table}[t]
\centering
\caption{Inference hyperparameters for \textbf{en$\rightarrow$zh}.}
\label{tab:final_en-zh-hypers}
\begin{tabular}{c|c|c|c|c}
\hline
Latency & MUD     & VPT      & MSD  & MCS \\ \hline
low            & 0.1  & 0.5 & 0.5  & 5    \\
high           & 0.1  & 0.5 & 1.5  & 7    \\ \hline
\end{tabular}
\end{table}

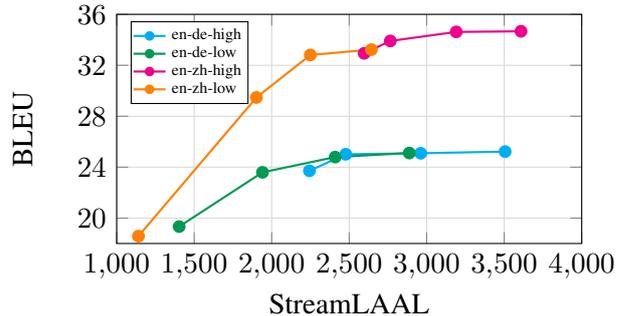
\begin{figure}[htbp!]
  \centering
  \begin{tikzpicture}
    \begin{axis}[
      width=\columnwidth,              
      height=0.6\columnwidth,         
      xlabel={StreamLAAL},
      ylabel={BLEU},
      xmin=1000, xmax=4000,
      ymin=18, ymax=36,
      xtick={1000,1500, 2000, 2500, 3000, 3500, 4000},
      ytick={20,24,28,32,36},
      grid=major,
      grid style={gray!30},
      legend pos=north west,
      legend cell align=left,
      legend style={font=\tiny, cells={anchor=west}, inner sep=0.5pt, row sep=-3pt},
      legend image post style={scale=0.7}
    ]
      \addplot[cyan,   mark=*,       thick] coordinates { (2243.48,23.71) (2477.67,25.01) (2961.48, 25.09) (3505.81, 25.22)  };
      \addplot[ForestGreen,    mark=*,  thick] coordinates { (1402.86,19.33) (1940.58,23.59) (2409.50,24.79) (2887.78,25.11)};
        \addplot[magenta,    mark=*,  thick] coordinates { (2596.57,32.94) (2765.39,33.91) (3190.68,34.62) (3608.24,34.67)};
      \addplot[orange,    mark=*,  thick] coordinates { (1140.47,18.58) (1901.35,29.47) (2249.32,32.81) (2643.12,33.23)};
      \legend{en-de-high, en-de-low, en-zh-high, en-zh-low}
    \end{axis}
  \end{tikzpicture}
  \caption{BLEU score plotted against StreamLAAL on the \textbf{en$\rightarrow$de} and \textbf{en$\rightarrow$zh} language pairs for minimum chunk sizes of 1, 3, 5, 7.}
  \label{fig:chunkhyper}
\end{figure}

Using the optimal maximum unvoiced duration, voice probability threshold, and maximum segment duration from Tables \ref{tab:final_en-de-hypers} and \ref{tab:final_en-zh-hypers} of our previous search, we iteratively step through a minimum chunk size of 1, 3, 5, and 7. We report the results for the BLEU and StreamLAAL for each given chunk size for both language pairs at the low and high latency regimes in Figure \ref{fig:chunkhyper}. Our final selected minimum chunk size for each latency regime is reported in Tables \ref{tab:dehyper} and \ref{tab:zhhyper}. 

\subsection{Quality and Latency Results on Dev set}
We provide the quality and latency of our system in Table \ref{tab:comparisons} on the ACL 60/60 dev set for the en$\rightarrow$de and en$\rightarrow$zh language pairs.  For each language pair, we show the influence of model size on our system's BLEU score.  From the results in Table \ref{tab:comparisons}, we can see that increasing the model size from 4B to 12B can offer approximately a 2 BLEU increase.  We choose to submit the 12B Gemma 3 translation model version of our cascaded architecture to the IWSLT 2025 simultaneous track \cite{abdulmumin-etal-2025-findings}.  On the en$\rightarrow$de language pair, we achieve a BLEU of 24.64 and 27.83 on the low and high latency regimes.  Then on the en$\rightarrow$zh language pair, we achieve a BLEU of 34.07 and 37.23 on the low and high latency regimes.

\section{Conclusion}
In this paper, we provide Oregon State University's system, BeaverTalk,  designed for the low and high latency regimes of the en$\rightarrow$de and en$\rightarrow$zh language pairs as a part of the IWSLT 2025 simultaneous track.  Our system consists of a cascaded architecture composed of a VAD speech segmenter, a Whisper ASR model, and a Gemma 3 translation LLM using conversational prompting.  We provide an extensive inference hyperparameter search for our system and demonstrate its performance utilizing a 4B and 12B translation LLM.  Our final submitted model, composed of the 12B translation LLM, demonstrates strong results on the en$\rightarrow$de and en$\rightarrow$zh language pairs for both the low and high latency categories.

\section*{Acknowledgements}
This research was supported, in part, by the National Science Foundation grants 2223483 and 2223484.

\bibliography{custom}

\appendix


\end{document}